%% file: main.tex
\documentclass[sigconf]{acmart}
\renewcommand\footnotetextcopyrightpermission[1]{} 
\setcopyright{none}

\usepackage{booktabs} 
\usepackage[utf8]{inputenc} 
\usepackage[T1]{fontenc}    
\usepackage{hyperref}       
\usepackage{url}            
\usepackage{booktabs}       
\usepackage{amsfonts}       
\usepackage{nicefrac}       
\usepackage{microtype}      
\usepackage{amsthm}
\usepackage{times}
\usepackage{epsfig}
\usepackage{graphicx}
\usepackage{amsmath}
\usepackage{amssymb}
\usepackage{extarrows}
\usepackage{wrapfig}
\usepackage{bm}
\usepackage{subcaption}
\usepackage{todonotes}
\usepackage{tcolorbox}
\usepackage{tikz}
\usepackage{tikz-3dplot}
\usepackage{tikzscale}
\usetikzlibrary{arrows,shapes,chains,matrix,positioning}
\usetikzlibrary{scopes,decorations,shadows,backgrounds,fit}
\usetikzlibrary{decorations.pathreplacing,calc,3d,quotes}
\usepackage{pgfmath}
\usepackage{threeparttable}
\usepackage{color}
\usepackage{array}
\usepackage{mdwmath}
\usepackage{mdwtab}
\usepackage{eqparbox}
\usepackage{fixltx2e}
\usepackage{stfloats}
\usepackage{diagbox}
\usepackage{verbatim}
\usepackage{multirow}
\usepackage{mathtools}
\usepackage{breqn}
\usepackage{algorithmic}
\usepackage[]{algorithm2e}
\usepackage{listings}
\usepackage{color} 
\definecolor{mygreen}{RGB}{28,172,0} 
\definecolor{mylilas}{RGB}{170,55,241}
\def\etal{{\it et al. }}

\def\ourmethod{NSGA-Net}

\theoremstyle{definition}

\theoremstyle{remark}

\newcommand{\tworows}[2]{\begin{tabular}{@{}c@{}} #1 \\ #2 \end{tabular}}
\def\*#1{\mathbf{#1}}
\usepackage[symbol]{footmisc}

\definecolor{nsganet}{rgb}{0.4, 0.7607843137254902, 0.6470588235294118}
\definecolor{nasnet}{rgb}{0.9882352941176471, 0.5529411764705883, 0.3843137254901961}
\definecolor{amoebanet}{rgb}{0.5529411764705883, 0.6274509803921569, 0.796078431372549}
\definecolor{darts}{rgb}{0.9058823529411765, 0.5411764705882353, 0.7647058823529411}
\definecolor{mygreen}{rgb}{0.53, 0.66, 0.42}
\definecolor{myyellow}{rgb}{1.0, 0.88, 0.21}
\definecolor{brilliantrose}{rgb}{1.0, 0.33, 0.64}

\def\ourmethod{NSGA-Net}
\def\sota{state-of-the-art}

\begin{document}
\title[\ourmethod{}: Neural Architecture Search using EMO]{\ourmethod{}: Neural Architecture Search using Multi-Objective Genetic Algorithm}

\author{Zhichao Lu, Ian Whalen, Vishnu Boddeti, Yashesh Dhebar, \\
Kalyanmoy Deb, Erik Goodman and Wolfgang Banzhaf}
\affiliation{%
  \institution{Michigan State University}
  \streetaddress{XXXXX}
  \city{East Lansing} 
  \state{Michigan} 
  \postcode{48824}
}
\email{{luzhicha, whalenia, vishnu, dhebarya, kdeb, goodman, banzhafw}@msu.edu}

\input{abstract.tex}
\maketitle
\input{introduction.tex}
\input{related-work.tex}
\input{approach.tex}
\input{experiments.tex}
\input{conclusion.tex}
\input{appendix.tex}

\bibliographystyle{ACM-Reference-Format}
\bibliography{egbib} 

\end{document}

%% file: abstract.tex
\begin{abstract}
    This paper introduces \ourmethod{} -- an evolutionary approach for neural architecture search (NAS). \ourmethod{} is designed with three goals in mind: (1) a procedure considering multiple and conflicting objectives, (2) an efficient procedure balancing exploration and exploitation of the space of potential neural network architectures, and (3) a procedure finding a diverse set of trade-off network architectures achieved in a single run. \ourmethod{} is a population-based search algorithm that explores a space of potential neural network architectures in three steps, namely, a population \emph{initialization} step that is based on prior-knowledge from hand-crafted architectures, an \emph{exploration} step comprising crossover and mutation of architectures, and finally an \emph{exploitation} step that utilizes the hidden useful knowledge stored in the entire history of evaluated neural architectures in the form of a Bayesian Network. Experimental results suggest that combining the dual objectives of minimizing an error metric and computational complexity, as measured by FLOPs, allows \ourmethod{} to find competitive neural architectures. Moreover, \ourmethod{} achieves error rate on the CIFAR-10 dataset on par with other state-of-the-art NAS methods while using orders of magnitude less computational resources. These results are encouraging and shows the promise to further use of EC methods in various deep-learning paradigms.
\end{abstract}

%% file: introduction.tex
\section{Introduction}
Deep convolutional neural networks have been overwhelmingly successful in a variety of image analysis tasks. One of the key driving forces behind this success is the introduction of many CNN architectures, such as AlexNet \citep{krizhevsky2012alexnet}, VGG \citep{simonyan2015vgg}, GoogLeNet \citep{szegedy2015inception}, ResNet \citep{he2016resnet}, DenseNet \citep{huang2017densenet} etc. in the context of image classification. Concurrently, network designs such as MobileNet \citep{howard2017mobilenets}, XNOR-Net \citep{rastegari2016xnor}, BinaryNets \citep{courbariaux2016binarized}, LBCNN \citep{juefei2017local} etc. have been developed with the goal of enabling real-world deployment of high performance models on resource constrained devices. These developments are the fruits of years of painstaking efforts and human ingenuity.

Neural architecture search (NAS) methods, on the other hand, seek to automate the process of designing network architectures. State-of-the-art reinforcement learning (RL) methods like \citep{real2017largescale} and \citep{nasnet2018} are inefficient in their use of their search space and require 3,150 and 2,000 GPU days, respectively. Gradient-based methods like \citep{liu2018darts} focus on the single objective of minimizing an error metric on a task and cannot be easily adapted to handle multiple conflicting objectives. Furthermore, most \sota{} approaches search over a single computation block, similar to an Inception block \citep{szegedy2015inception}, and repeat it as many times as necessary to form a complete network. 

\begin{figure*}[hbt]
\centering
\includegraphics[width=0.9\textwidth]{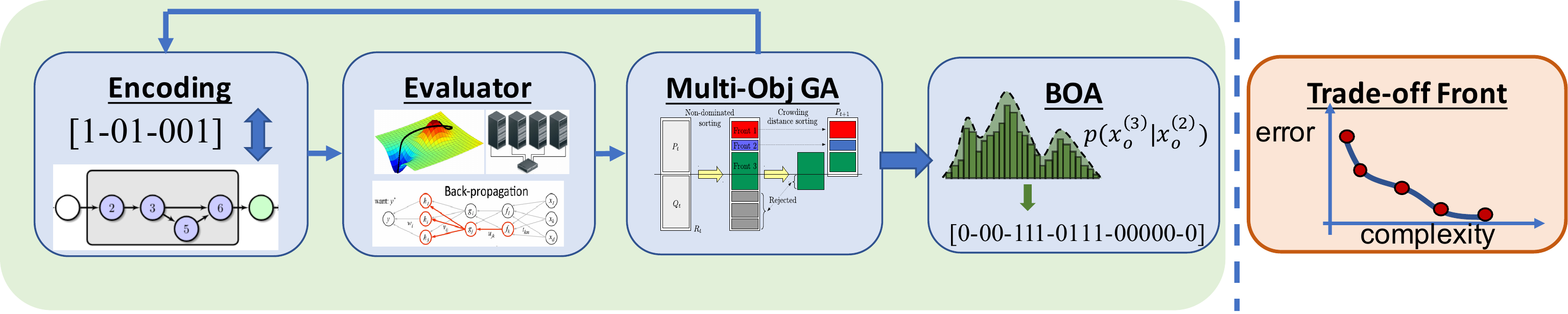}
\caption{Overview of the stages of \ourmethod{}. Networks are represented as bit strings, trained through gradient descent, ranking and selection by NSGA-II, search history exploitation through BOA. Output is a set of networks that span a range of complexity and error objectives.}
\label{fig:overview}
\end{figure*}

In this paper, we present \ourmethod{}, a multi-objective genetic algorithm for NAS to address the aforementioned limitations of current approaches. A pictorial overview of \ourmethod{} is provided in Figure~\ref{fig:overview}. The salient features of \ourmethod{} are,

\begin{enumerate}
\item \textbf{Multi-Objective Optimization}: Real-world deployment of NAS models demands small-sized networks, in addition the models being accurate. For instance, we seek to maximize performance on compute devices that are often constrained by hardware resources in terms of power consumption, available memory, available FLOPs, and latency constraints, to name a few. \ourmethod{} is explicitly designed to optimize such competing objectives.

\item \textbf{Flexible Architecture Search Space}: The search space for most existing methods is restricted to a block that is repeated as many times as desired. In contrast, \ourmethod{} searches over the entire structure of the network. This scheme overcomes the limitations inherent in repeating the same computation block throughout an entire network, namely, that a single block may not be optimal for every application and it is desirable to allow NAS to discover architectures with different blocks in different parts of the network.

\item \textbf{Non-Dominated Sorting}: The core component of \ourmethod{} is the Non-Dominated Sorting Genetic Algorithm II (NSGA-II) \citep{deb2000fast}, a multi-objective optimization algorithm that has been successfully employed for solving a variety of multi-objective problems \citep{tapia2007applications, pedersen2006multi}. Here, we leverage its ability to maintain a diverse trade-off frontier between multiple conflicting objectives, thereby resulting in a more effective and efficient exploration of the search space.

\item \textbf{Efficient Recombination}: In contrast to \sota{} evolution-based NAS methods \citep{real2017largescale,real2018amoebanet} in which only mutation is used, we employ crossover (in addition to mutation) to combine networks with desirable qualities across multiple objectives from the diverse frontier of solutions. 

\item \textbf{Bayesian Learning}: We construct and employ a Bayesian Network inspired by the Bayesian Optimization Algorithm (BOA) \citep{pelikan1999boa} to fully utilize the promising solutions present in our search history archive and the inherent correlations between the layers of the network architecture.
\end{enumerate}

We demonstrate the efficacy of \ourmethod{} on the CIFAR10 \cite{cifar10} image classification task by minimizing two objectives: classification error and computational complexity. Here, computational complexity is defined by the number of floating-point operations (FLOPs) that a network carries out during a forward pass. Experimentally, we observe that \ourmethod{} can find a set of network architectures containing solutions that are significantly better than hand-crafted methods in both objectives, while being competitive with single objective state-of-the-art NAS approaches. Furthermore, by fully utilizing a population of networks through recombination and utilization of the search history, \ourmethod{} explores the search space efficiently and requires less computational time for search than other competing methods. The implementation of \ourmethod{} is available  \hyperlink{https://github.com/ianwhale/nsga-net}{here\footnote{\textcolor{brilliantrose}{https://github.com/ianwhale/nsga-net}}}.

%% file: related-work.tex
\vspace{-1.5mm}
\section{Related Work}

Recent research efforts in NAS have produced a plethora of methods to automate the design of networks. Broadly speaking, these methods can be divided into evolutionary algorithm (EA) and reinforcement learning (RL) based approaches -- with a few methods falling outside these two categories. The main motivation of EA methods is to treat structuring a network as a combinatorial optimization problem. EAs operate with a population that makes small changes (mutation) and mixes parts (crossover) of solutions selected by consideration of multiple objectives to guide its search toward the optimal solutions. RL, on the other hand, views the construction of a network as a decision process. Usually, an agent is trained to optimally choose the pieces of a network in a particular order. We briefly review a few existing methods here.

\vspace{3pt}
\noindent\textbf{Reinforcement Learning:} $Q$-learning \citep{watkins1989qlearning} is a widely popular value iteration method used for RL. The MetaQNN method \citep{baker2017metaqnn} employs an $\epsilon$-greedy $Q$-learning strategy with experience replay to search connections between convolution, pooling, and fully connected layers, and the operations carried out inside the layers. Zhong et al. \citep{zhong2017blockqnn} extended this idea with the BlockQNN method. BlockQNN searches the design of a computational block with the same $Q$-learning approach. The block is then repeated to construct a network. This method allows for a much more general network and achieves better results than its predecessor on CIFAR-10 \citep{cifar10}.

A policy gradient method seeks to approximate some not differentiable reward function to train a model that requires parameter gradients, like a neural network architecture. \citet{zoph2016} first applied this method in architecture search to train a recurrent neural network controller that constructs networks. The original method in \citep{zoph2016} uses the controller to generate the entire network at once. This contrasts from its successor, NASNet \citep{nasnet2018}, which designs a convolutional and pooling block that is repeated to construct a network. NASNet outperforms its predecessor and produces a network achieving state-of-the-art error rate on CIFAR-10. \ourmethod{} differs from RL methods by using more than one selection criteria. More specifically, networks are selected for their accuracy on a task, rather than an approximation of accuracy, along with computational complexity. Furthermore, the most successful RL methods search only a computational block that is repeated to create a network, \ourmethod{} allows for search across computational blocks and combinations of blocks. \citet{hsu2018monas} extends the NASNet approach to multi-objective domain to optimize multiple linear combinations of accuracy and energy consumption criteria for different scalarization parameters. However, multiple generative applications of a scalarized objectives was shown to be not as efficient as simultaneous approaches \citep{deb2000fast}.

\vspace{3pt}
\noindent\textbf{Evolutionary Algorithms:}  Designing neural networks through evolution, or \textit{neuroevolution}, has been a topic of interest for some time, first showing popular success in 2002 with the advent of the neuroevolution of augmenting topologies (NEAT) algorithm \citep{stanley2002neat}. In its original form, NEAT only performs well on comparatively small networks. \citet{miikulainen2017codeepneat} attempted to extend NEAT to deep networks with CoDeepNEAT using a co-evolutionary approach that achieved limited results on the CIFAR-10 dataset. CoDeepNEAT does, however, produce state-of-the-art results in the Omniglot multi-task learning domain \citep{liang2018codeepneatomniglot}. 

\citet{real2017largescale} introduced perhaps the first truly large scale application of a simple evolutionary algorithm. The extension of this method presented in \citep{real2018amoebanet}, called AmoebaNet, provided the first large scale comparison of EC and RL methods. Their simple EA searches over the same space as NASNet \citep{nasnet2018} and has shown to have a faster convergence to an accurate network when compared to RL and random search. Furthermore, AmoebaNet produced one of the best state-of-the-art results on CIFAR-10 data-set.

Conceptually, \ourmethod{} is closest to the Genetic CNN \citep{genetic_cnn} algorithm. It uses a binary encoding that corresponds to connections in convolutional blocks. In \ourmethod{}, we augment the original encoding and genetic operations by (1) adding an extra bit for a residual connection, and (2) introducing phase-wise crossover. We also introduce a multi-objective based selection scheme. Moreover, we also diverge from Genetic CNN by incorporating a Bayesian network in our search to fully utilize past population history as learned knowledge.

Evolutionary multi-objective optimization (EMO) approaches have been scarcely used for NAS. \citet{kim2017nemo} presents an algorithm utilizing NSGA-II \citep{deb2000fast}, however their method only searches over hyper-parameters and a small fixed set of architectures. The evolutionary method shown in \citep{elsken2018lemonade} uses weight sharing through network morphisms \citep{wei2016morhpisms} and approximate morphisms as mutations and uses a biased sampling to select for novelty from the objective space rather than a principled selection scheme, like NSGA-II \citep{deb2000fast}. Network morphisms allow for a network to be ``widened" or ``deepened" in a manner that maintains functional equivalence. For architecture search, this allows for easy parameter sharing after a perturbation in a network's architecture.

\vspace{3pt}
\noindent\textbf{Other Methods:} Methods that do not subscribe to either an EA or RL paradigm have also shown success in architecture search. \citet{liu2017progressive} presents a method that progressively expands networks from simple cells and only trains the best $K$ networks that are predicted to be promising by a RNN meta-model of the encoding space. \citet{dong2018ppp-net} extended this method to use a multi-objective approach, selected the $K$ networks based on their Pareto-optimality when compared to other networks. \citet{luo2018nao} also presents a meta-modeling approach that generates models with state-of-the-art accuracy. This approach may be ad-hoc as no analysis is presented on how the progressive search affects the trade-off frontier. \citet{elsken2018simple} use a simple hill climbing method along with a network morphism \citep{wei2016morhpisms} approach to optimize network architectures quickly on limited resources. \citet{chen2018rl_ea_nas} combine the ideas of RL and EA. A population of networks is maintained and are selected for mutation with tournament selection \citep{goldberg1991tournament}. A recurrent network is used as a controller to learn an effective strategy to apply mutations to networks. Networks are then trained and the worst performing network in the population is replaced. This approach generates state of the art results for the ImageNet classification task. \citet{chen2018randomsearch} presented an augmented random search approach to optimize networks for a semantic segmentation application.  \citet{kandasamy2018bayesian} presents a Gaussian process based approach to optimize network architectures, viewing the process through a Bayesian optimization lens.

%% file: approach.tex

\begin{figure*}[t]
    \centering
    \includegraphics[width=0.92\textwidth{}]{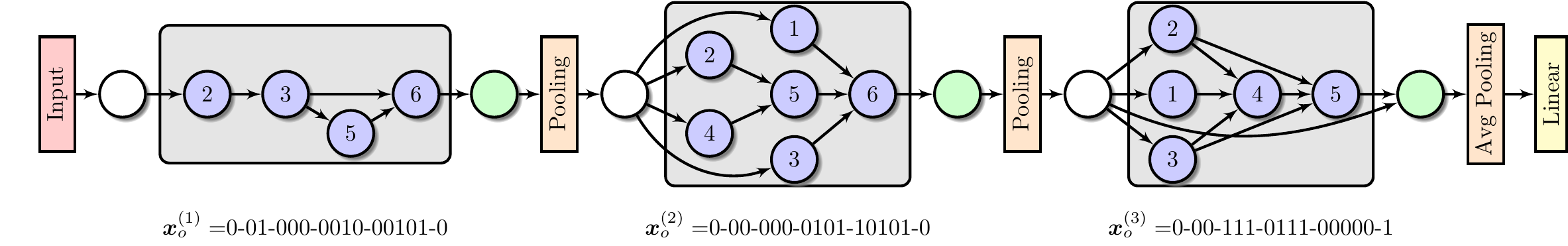}
    \caption{\textbf{Encoding:} Illustration of a classification network encoded by $\bf{x} = \bf{x_o}$, where $\bm{x}_o$ is the operations at a phase (gray boxes, each with a possible maximum of 6 nodes). In this example the spatial resolution changes (orange boxes that connect the phases) are fixed based on prior knowledge of successful approaches. The phases are described by the bit string $\mathbf{x_o}$ which is formatted for readability above. The bits are grouped by dashes to describe what node they control. See Section~\ref{sec:encoding} for detailed description of the encoding schemes.}
    \label{fig:explicit_encoding}
\end{figure*}

\section{Proposed Approach}\label{sec:approach}
Compute devices are often constrained by hardware resources in terms of their power consumption, available memory, available FLOPs, and latency constraints. Hence, real-world design of DNNs are required to balance these multiple objectives (e.g., predictive performance and computational complexity). Often, when multiple design criteria are considered simultaneously, there may not exist a single solution that performs optimally in all desired criteria, especially with competing objectives. Under such circumstances, a set of solutions that provide representative trade-off information between the objectives is more desirable. This enables a practitioner to analyze the importance of each criterion, depending on the application, and to choose an appropriate solution on the trade-off frontier for implementation. We propose \ourmethod{}, a genetic algorithm based architecture search method to automatically generate a set of DNN architectures that approximate the Pareto-front between performance and complexity on an image classification task. The rest of this section describes the encoding scheme, and main components of \ourmethod{} in detail.

\begin{figure*}
    \centering
    \includegraphics[width=0.85\textwidth]{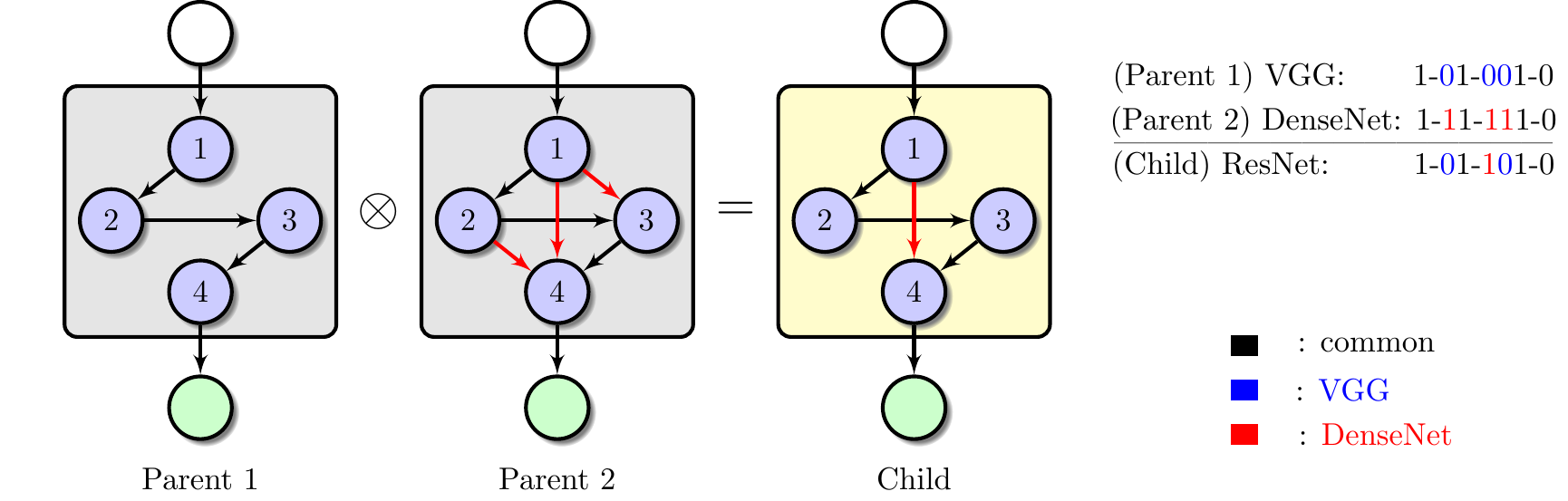}
    \caption{\textbf{Crossover Example:} A crossover (denoted by $\otimes$) of a VGG-like structure with a DenseNet-like structure may result in a ResNet-like network. In the figure, {\color{red}{red}} and {\color{blue}{blue}} denotes connections that are unique to VGG and DenseNet respectively, and black shows the connections that are common to both parents. All black bits are retained in the final child encoding, and only the bits that are not common between the parents can potentially be selected at random from one of the parent.}
    \label{fig:crossover}
\end{figure*}

\subsection{Encoding}\label{sec:encoding}
Genetic algorithms, like any other biologically inspired search methods, often do not directly operate on {\em phenotypes}. From the biological perspective, we may view the DNN architecture as a phenotype, and the representation it is mapped from as its {\em genotype}. As in the natural world, genetic operations like crossover and mutation are only carried out in the genotype space; such is the case in \ourmethod{} as well. We refer to the interface between the genotype and the phenotype as \textit{encoding} in this paper. 

Most existing CNN architectures can be viewed as a composition of computational blocks that define the layer-wise computation (e.g. ResNet blocks \citep{he2016resnet}, DenseNet block \citep{huang2017densenet}, and Inception block \citep{szegedy2015inception}, etc.) and a scheme that specifies the spatial resolution changes. For example, down-sampling is often used after computational blocks to reduce the spatial resolution of information going into the next computational blocks in image classification DNNs. In \ourmethod{}, each computational block, referred to as a \textit{phase}, is encoded using the method presented by \citet{genetic_cnn}, with the small change of adding a bit to represent a skip connection that forwards the input information directly to the output bypassing the entire block. And we name it as the \textit{Operation Encoding} $\*{x_o}$ in this study.

\vspace{3pt}
\noindent\textbf{Operation Encoding $\*{x_o}$}: Unlike most of the hand-crafted and NAS generated architectures, we do not repeat the same phase (computational block) to construct a network. Instead, the operations of a network are encoded by $\*{x_o} = \left(\*{x_o^{(1)}}, \*{x_o^{(2)}}, \dots, \*{x_o^{(n_p)}}\right)$ where $n_p$ is the number of phases. Each $\*{x_o^{(i)}}$ encodes a directed acyclic graph consisting of $n_o$ number of nodes that describes the operation within a phase using a binary string. Here, a \textit{node} is a basic computational unit, which can be a single operation like convolution, pooling, batch-normalization \citep{batch-norm} or a sequence of operations. This encoding scheme offers a compact representation of the network architectures in genotype space, yet is flexible enough that many of the computational blocks in hand-crafted networks can be encoded, e.g. VGG \citep{simonyan2015vgg}, ResNet \citep{he2016resnet} and DenseNet \citep{huang2017densenet}. Figure~\ref{fig:explicit_encoding} and Figure~\ref{fig:crossover} shows examples of the operation encoding.

\vspace{3pt}
\noindent\textbf{Search Space}: 
With a pre-determined scheme of spatial resolution reduction (similarly in \citep{nasnet2018,real2018amoebanet,liu2018darts}), the total search space in the genotype space is governed by our operation encoding $\*{x_o}$:
\begin{align*}
     \Omega_{\*{x}} = \Omega_{\*{x_o}} = n_p \times 2^{n_o(n_o - 1)/2 + 1}
\end{align*}
\noindent where $n_p$ is the number of phases (computational blocks), and $n_o$ is the number of nodes (basic computational units) in each phase. However, for computationally tractability, we constrain the search space such that each node in a phase carries the same sequence of operations, i.e. a $3\times3$ convolution followed by batch-normalization \citep{batch-norm} and ReLU.

It is worth noting that, as a result of nodes in each phase having identical operations, the encoding between genotype and phenotype is a many-to-one mapping. Given the prohibitive computational expense required to train each network architecture before its performance can be assessed, it is essential to avoid evaluating genomes that decode to the same architecture. We develop an algorithm to quickly and approximately identify these duplicate genomes (see Appendix for details).

\subsection{Search Procedure}
\ourmethod{} is an iterative process in which initial solutions are made gradually better as a group, called a \textit{population}. In every iteration, the same number of offspring (new network architectures) are generated from parents selected from the population. Each population member (including both parents and offspring) compete for both survival and reproduction (becoming a parent) in the next iteration. The initial population may be generated randomly or guided by prior-knowledge (e.g. seeding the hand-crafted network architectures into the initial population). Following initialization, the overall \ourmethod{} search proceeds in two sequential stages, an \emph{exploration} and \emph{exploitation}. 

\vspace{3pt}
\noindent\textbf{Exploration}:
The goal of this stage is to discover diverse ways of connecting nodes to form a phase (computational block). Genetic operations, crossover and mutation, offer an effective mean to realize this goal.

\vspace{3pt}
\noindent\textbf{Crossover:} The {\em implicit\/} parallelism of population-based search approaches can be unlocked when the population members can effectively share (through crossover) building-blocks \citep{holland-book}. In the context of NAS, a phase or the sub-structure of a phase can be viewed as a building-block. We design a homogeneous crossover operator, which takes two selected population members as parents, to create offspring (new network architectures) by inheriting and recombining the building-blocks from parents. The main idea of this crossover operator is to 1) preserve the common building-blocks shared between both parents by inheriting the common bits from both parents' binary bit-strings; 2) maintain, relatively, the same complexity between the parents and their offspring by restricting the number of ``1" bits in the offspring's bit-string to lie between the number of ``1" bits in both parents. The proposed crossover allows selected architectures (parents) to effectively exchange phases or sub-structures within a phase. An example of the crossover operator is provided in Figure~\ref{fig:crossover}.

\vspace{3pt}
\noindent\textbf{Mutation:} To enhance the diversity (having different network architectures) of the population and the ability to escape from local optima, we use a bit-flipping mutation operator, which is commonly used in binary-coded genetic algorithms. Due to the nature of our encoding, a one bit flip in the genotype space could potentially create a completely different architecture in the phenotype space. Hence, we restrict the number of bits that can be flipped to be at most one for each mutation operation. As a result, only one of the phase architectures can be mutated at one time. 

\begin{figure*}[t]
\centering
\includegraphics[width=0.75\textwidth]{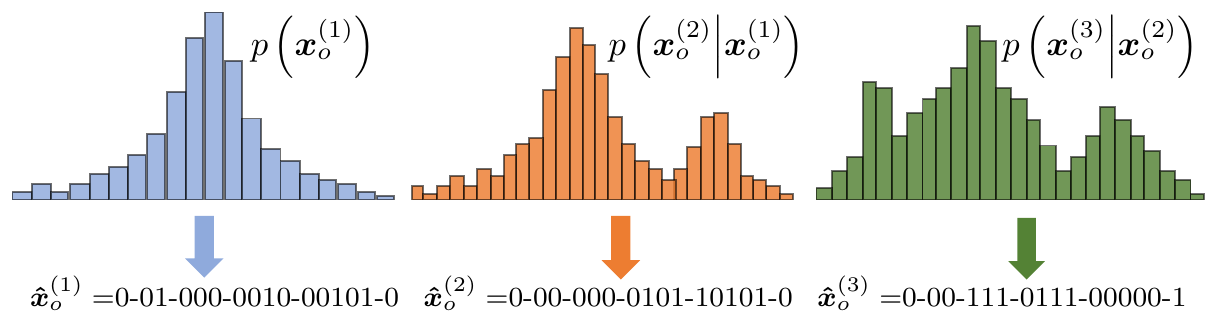}
\caption{\textbf{Exploitation:} Sampling from the Bayesian Network (BN) constructed by \ourmethod{}. The histograms represent estimates of the conditional distributions between the network structure between the phases explored during the exploration step and updated during the exploitation step (i.e., using the population archive). During exploitation, networks are constructed by sampling phases from the BN. Fig.~\ref{fig:explicit_encoding} shows the architectures that the sampled bit strings, $\{\bm{\hat{x}}_o^{(1)},\bm{\hat{x}}_o^{(2)},\bm{\hat{x}}_o^{(3)} \}$  decode to. \label{fig:exploitation}}
\end{figure*}

\vspace{3pt}
\noindent\textbf{Exploitation}: The exploitation stage follows the exploration stage in \ourmethod{}. The goal of this stage is to exploit and reinforce the patterns commonly shared among the past successful architectures explored in the previous stage. The exploitation step in \ourmethod{} is heavily inspired by the Bayesian Optimization Algorithm (BOA) \citep{pelikan1999boa} which is explicitly designed for problems with inherent correlations between the optimization variables. In the context of our NAS encoding, this translates to correlations in the blocks and paths across the different phases. Exploitation uses past information across all networks evaluated to guide the final part of the search. More specifically, say we have a network with three phases, namely $\*{x_o^{(1)}}$, $\*{x_o^{(2)}}$, and $\*{x_o^{(3)}}$. We would like to know the relationship of the three phases. For this purpose, we construct a Bayesian Network (BN) relating these variables, modeling the probability of networks beginning with a particular phase $\*{x_o^{(1)}}$, the probability that $\*{x_o^{(2)}}$ follows $\*{x_o^{(1)}}$, and the probability that $\*{x_o^{(3)}}$ follows $\*{x_o^{(2)}}$. In other words, we estimate the distributions $p\left(\*{x_o^{(1)}}\right)$, $p\left(\*{x_o^{(2)}} | \*{x_o^{(1)}}\right)$, and $p\left(\*{x_o^{(3)}} | \*{x_o^{(2)}}\right)$ by using the population history, and update these estimates during the exploitation process. New offspring solutions are created by sampling from this BN. Figure~\ref{fig:exploitation} shows a pictorial depiction of this process.

%% file: experiments.tex
\section{Experiments}
In this section, we explain the experimental setup and implementation details of \ourmethod{}, followed by the empirical results to demonstrate the efficacy of \ourmethod{} to automate the NAS process on image classification task.

\subsection{Performance Metrics}
We consider two objectives to guide \ourmethod{} based NAS, namely, classification error and computational complexity. A number of metrics can serve as proxies for computational complexity: number of active nodes, number of active connections between the nodes, number of parameters, inference time and number of floating-point operations (FLOPs) needed to execute the forward pass of a given network. Our initial experiments considered each of these different metrics. We concluded from extensive experimentation that inference time cannot be estimated reliably due differences and
inconsistencies in computing environment, GPU manufacturer, temperature, etc. Similarly, the number of parameters, active connections or active nodes only relate to one aspect of computational complexity. In contrast, we found an estimate of FLOPs to be a more accurate and reliable proxy for network complexity. See Appendix for more details. Therefore, classification error and FLOPs serve as the twin objectives for selecting networks. 

For the purpose of quantitatively comparing different multi - objective search methods or different configuration setups of \ourmethod{}, we use the hypervolume (HV) performance metric, which calculates the dominated area (hypervolume in the general case) from the a set of solutions (network architectures) to a reference point which is usually an estimate of the nadir point---a vector concatenating worst objective values of the Pareto-frontier. It has been proven that the maximum HV can only be achieved when all solutions are on the Pareto-frontier \citep{hypervolume}. Hence, the higher the HV measures, the better solutions that are being found in terms of both objectives. 

\subsection{Implementation Details}

\vspace{3pt}
\textbf{Dataset:} We consider the CIFAR-10 \citep{cifar10} dataset for our classification task. We split the original training set (80\%-20\%) to create our training and validation sets for architecture search. The original CIFAR-10 testing set is only utilized at the conclusion of the search to obtain the test accuracy for the models on the final trade-off front. 

\vspace{3pt}
\textbf{\ourmethod{} hyper-parameters:} We set the number of phases $n_p$ to three and the number of nodes in each phase $n_o$ to six. We also fix the spatial resolution changes scheme similarly as in \cite{nasnet2018}, in which a max-pooling with stride $2$ is placed after the first and the second phase, and a global average pooling layer after the last phase. The initial population is generated by uniform random sampling. The probabilities of crossover and mutation operations are set at 0.9 and 0.02 respectively. The population size is 40 and the number of generations is 20 for the exploration stage. And another ten generations for exploitation. Hence, a total of 1,200 network architectures are searched by \ourmethod{}. 

\vspace{3pt}
\textbf{Network training during searching:} During architecture search, we limit the number of filters (channels) in any node to 16 for each one of the generated network architecture. We then train them on our training set using standard stochastic gradient descent (SGD) back-propagation algorithm and a cosine annealing learning rate schedule \citep{cosine}. Our initial learning rate is 0.025 and we train for 25 epochs, which takes about 9 minutes on a NVIDIA 1080Ti GPU implementation in PyTorch \citep{paszke2017automatic}. Then the classification error is measured on our validation set. 

\subsection{Architecture Validation} \label{sec:extrapolation}
For comparing with other single-objective NAS methods, we adopt the training procedure used in \citep{liu2018darts} and a quick summary is given as follows. 

We extend the number of epochs to 600 with a batch-size of 96 to train the final selected models (could be the entire trade-off frontier architectures or a particular one chosen by the decision-maker). We also incorporate a data pre-processing technique \emph{cutout} \citep{cutout}, and a regularization technique \emph{scheduled path dropout} introduced in \citep{nasnet2018}. In addition, to further improve the training process, an auxiliary head classifier is appended to the architecture at approximately 2/3 depth (right after the second resolution-reduction operation). The loss from this auxiliary head classifier, scaled by a constant factor 0.4, is aggregated with the loss from the original architecture before back-propagation during training. Other hyper-parameters related to the back-propagation training remain the same as during the architecture search. 

For the fairness of the comparison among various NAS methods, we incorporate the NASNet-A cell \citep{nasnet2018}, the AmoebaNet-A cell \citep{real2018amoebanet} and the DARTS(second order) cell \citep{liu2018darts} into our training procedures and report their results under the same settings as \ourmethod{} found architectures.

\subsection{Results Analysis}
\begin{figure}[!htbp]
\centering
\includegraphics[width=0.43\textwidth]{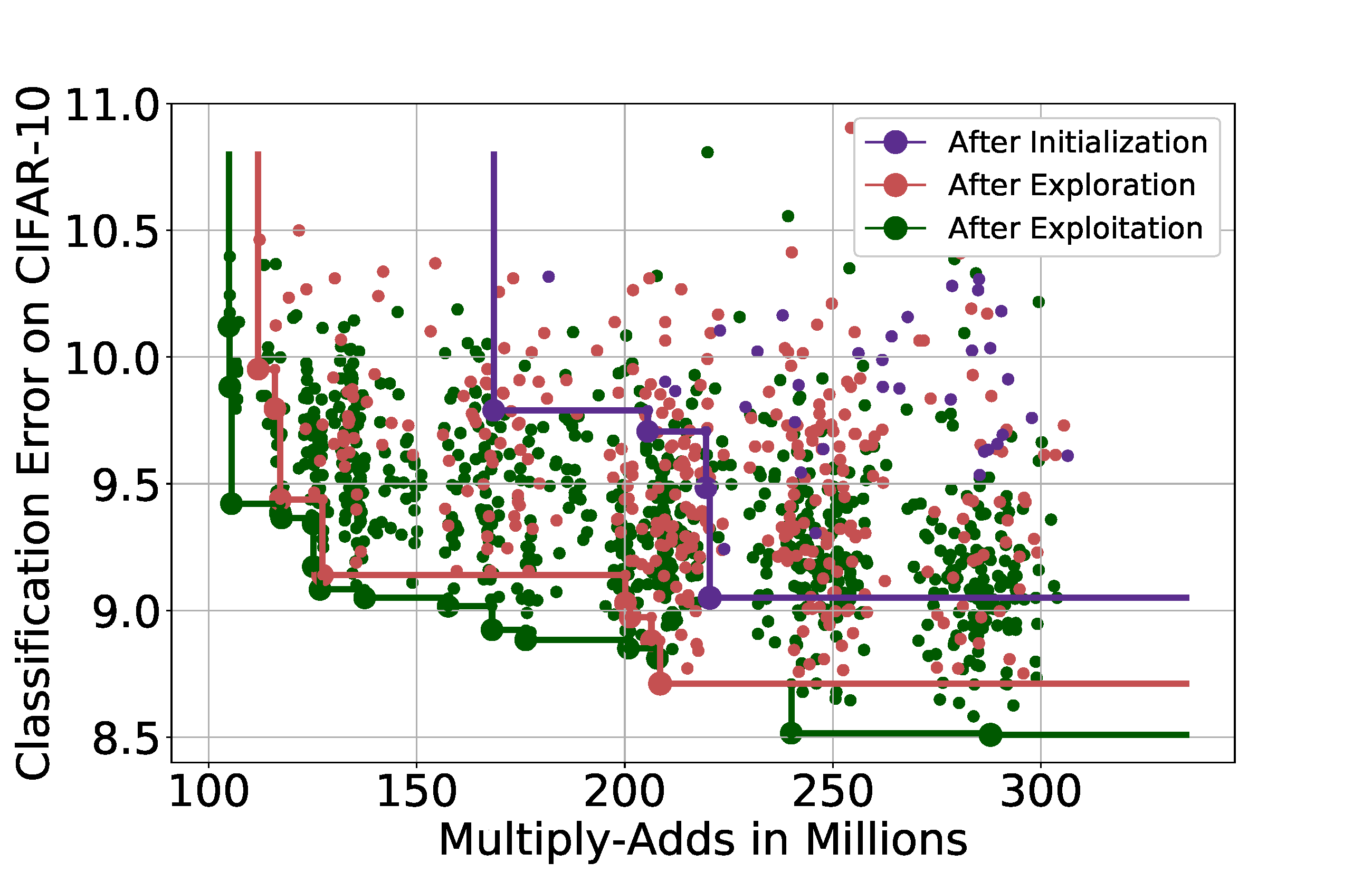}
\caption{Progression of trade-off frontiers after each stage of \ourmethod.}\label{fig:progress}
\end{figure}

\begin{figure}[!htbp]
\centering
\includegraphics[width=0.41\textwidth]{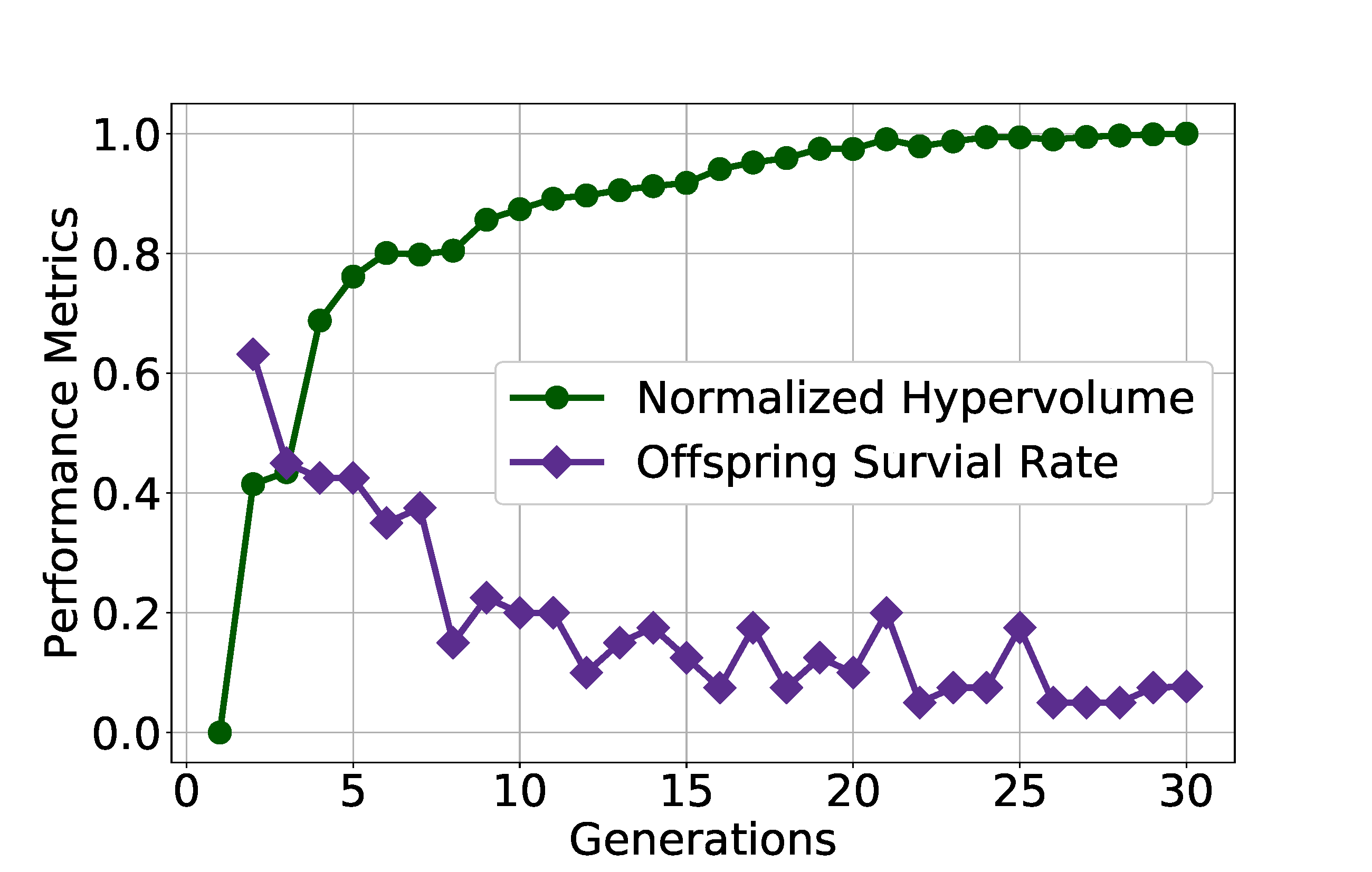}
\caption{Generational normalized hypervolume and survival rate of the offspring network architectures.}\label{fig:progress_meta}
\end{figure}

We first present the overall search progression of \ourmethod{} in the objective-space. Figure~\ref{fig:progress} shows the bi-objective frontiers obtained by \ourmethod{} through the various stages of the search, clearly showcasing a gradual improvement of the whole population. Figure~\ref{fig:progress_meta} shows two metrics: normalized HV and offspring survival rate, through the different generations of the population. The monotonic increase in the former suggests that a better set of trade-off network architectures have been found over the generations. The monotonic decrease in the latter metric suggests that, not surprisingly, it is increasingly difficult to create better offspring (than their parents). We can use a threshold on the offspring survival rate as a potential criterion to terminate the current stage of the search process and switch between the exploration and exploitation.

To compare the network architecture obtained from \ourmethod{} to other hand-crafted and search-generated architectures, we pick the network architectures with the lowest classification error from the final frontier (the dot in the lower right corner on the green curve in Figure~\ref{fig:progress}) and extrapolate (by following the setup as explained in Section~\ref{sec:extrapolation}) the network by increasing the number of filters of each node in the phases, and train with the entire official CIFAR-10 training set. The chosen network architecture, shown in Figure~\ref{fig:explicit_encoding}, results in 3.85\% classification error on the CIFAR-10 testing set with 3.34 Millions of parameters and 1290 MFLOPs. Table~\ref{tab:cifar10_compare} provides a summary that compares \ourmethod{} with other multi-objective NAS methods. Unfortunately, hypervolume comparisons between these multi-objective NAS methods are not feasible due to the following two reasons: 1) The entire trade-off frontiers obtained by the other multi-objective NAS methods are not reported and 2) different objectives were used to estimate the complexity of the architectures. Due to space limitation, the other architectures on the trade-off frontier found by \ourmethod{} are reported in Appendix. 

\begin{table}[t]
\caption{Multi-objective methods for CIFAR-10 (best accuracy for each method)}\label{tab:cifar10_compare}
\centering
\scalebox{0.85}{
\begin{tabular}{c c c c}
\toprule
Method & Error (\%) & Other Objective & Compute \\ \midrule 
PPP-Net \citep{dong2018ppp-net} & 4.36 & \tworows{FLOPs or}{Params or Inference Time} & Nvidia Titan X \\
& & & \\
MONAS \citep{hsu2018monas} & 4.34 & Power & Nvidia 1080Ti \\
& & & \\
\ourmethod{} & 3.85 & FLOPs & \tworows{Nvidia 1080Ti}{8 GPU Days} \\
 \bottomrule
\end{tabular}}
\end{table}

\begin{figure*}[!htbp]
\centering
\includegraphics[width=0.8\textwidth]{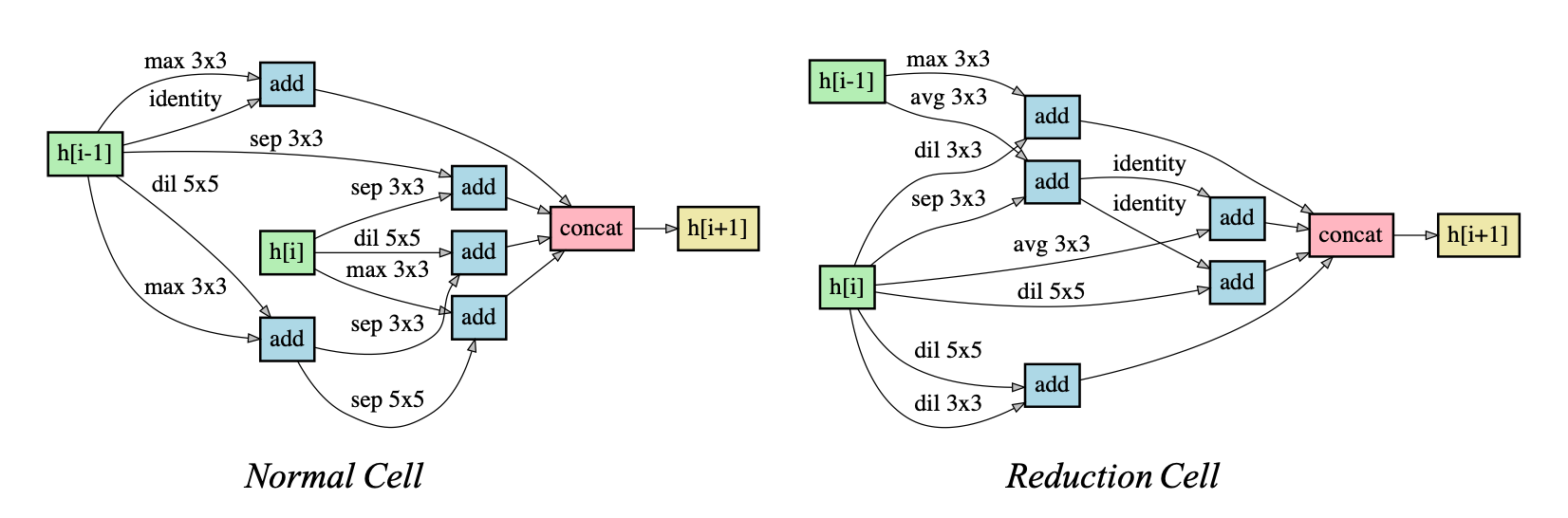}
\caption{Normal and reduction convolutional cell architectures found by \ourmethod{} applied to NASNet micro search space. The inputs (\textcolor{mygreen}{green}) are from previous cells' output (or input image). The output (\textcolor{myyellow}{yellow}) is the results of a concatenation operation across all resulting branches. Each edge (line with arrow) indicates an operation with operation name annotated above the line.} \label{fig:NASNet_cells}
\end{figure*}

\begin{table*}[!htbp]
\caption{Comparison of \ourmethod{} with baselines on CIFAR-10 image classification. In this table, the first block presents \sota{} architectures designed by human experts. The second block presents NAS methods that design the entire network. The last block presents NAS methods that design modular blocks which are repeatedly combined to form the final architecture. We use (N @ F) to indicate the configuration of each model, where N is the number of repetition and F is the number of filters right before classification. Results marked with $\dagger$ are obtained by training the corresponding architectures with our setup (refers to Section~\ref{sec:extrapolation} for details).}
\label{tab:cifar10}
\centering
\scalebox{1.0}{
\begin{tabular}{c@{\hspace{2mm}}l@{\hspace{1mm}}c@{\hspace{1mm}}c@{\hspace{2mm}}c@{\hspace{1mm}}c@{\hspace{1mm}} c@{\hspace{1mm}} c@{\hspace{2mm}} c@{\hspace{2mm}} c@{\hspace{3mm}}}
\toprule
 & Architectures & {Params}  & \multicolumn{1}{c}{Test Error} &{$\times$ \large{+}~~} & {Search Cost} & Search \\
 &  & {(M)}  & \multicolumn{1}{c}{(\%)} & {(M)} & {(GPU-days)} & Method

\\ \midrule
\multirow{4}{*}{\rotatebox{90}{}}
 & Wide ResNet \citep{zagoruyko2016wideresnet} & 36.5 & 4.17 & - & - & human experts \\
 & DenseNet-BC (k = 40) \citep{huang2017densenet} & 25.6 & 3.47 & - & - & human experts \\ \midrule\midrule
\multirow{3}{*}{\rotatebox{90}{}}
 & NAS \citep{zoph2016} & 7.1 & 4.47 & - & 3150 & RL\\
 & NAS + more filters\citep{zoph2016} & 37.4 & 3.65 & - & 3150 & RL\\
 & ENAS + macro search space \citep{enas2018} & 21.3 & 4.23 & - & 0.5 & RL + weight sharing\\
 & ENAS + macro search space + more channels \citep{enas2018} & 38.0 & 3.87 & - & 0.5 & RL + weight sharing\\
 \midrule
 \multirow{3}{*}{\rotatebox{90}{}}
 & \textbf{\ourmethod{} + macro search space} & \textbf{3.3} & \textbf{3.85} & \textbf{1290} & \textbf{8} & \textbf{evolution} \\
 \midrule\midrule
\multirow{3}{*}{\rotatebox{90}{}}
& DARTS second order + cutout \citep{liu2018darts} & 3.3 & 2.76 & - & 4 & gradient-based \\
 & DARTS second order (6 @ 576) + cutout \citep{liu2018darts} \footnote[2]{} & 3.3 & 2.76 & 547 & 4 & gradient-based \\
 & NASNet-A + cutout \citep{nasnet2018} & 3.3 & 2.65 & - & 2,000 & RL \\ 
 & NASNet-A (6 @ 660) + cutout \citep{nasnet2018} \footnote[2]{} & 3.2 & 2.91 & 532 & 2,000 & RL \\ 
 & ENAS + cutout \citep{enas2018} & 4.6 & 2.89 & - & 0.5 & RL + weight sharing\\ 
 & ENAS (6 @ 660) + cutout \citep{enas2018} \footnote[2]{} & 3.3 & 2.75 & 533 & 0.5 & RL + weight sharing\\ 
 & AmoebaNet-A \citep{real2018amoebanet} & 3.2 & 3.34 & - & 3,150 & evolution\\ 
 & AmoebaNet-A (6 @ 444) + cutout \citep{real2018amoebanet} \footnote[2]{} & 3.3 & 2.77 & 533 & 3,150 & evolution\\ 
 
 \midrule
 & \textbf{\ourmethod{} (6 @ 560) + cutout} & \textbf{3.3} & \textbf{2.75} & \textbf{535} & \textbf{4} & \textbf{evolution} \\
  & \textbf{\ourmethod{} (7 @ 1536) + cutout} & \textbf{26.8} & \textbf{2.50} & \textbf{4147} & \textbf{4} & \textbf{evolution} \\
 \bottomrule
\end{tabular}}
\end{table*}

The CIFAR-10 results comparing \sota{} CNN architectures from both human-designed and search-generated are presented in Table~\ref{tab:cifar10}. \ourmethod{} achieves comparable results with \sota{} architectures designed by human experts \citep{huang2017densenet} while having order of magnitude less parameters in the obtained network architecture. When compared with other \sota{} RL- and evolution-based NAS methods \citep{nasnet2018,real2018amoebanet}, \ourmethod{} achieves similar performance by using two and half orders of magnitude less computation resources (GPU-days). Even though \ourmethod{} falls short in search efficiency when compared to the gradient-based NAS method DARTS \citep{liu2018darts} despite a slight advantage in test error, it's worth noting that \ourmethod{} inherently delivers many other architectures from the trade-off frontier at no extra cost. The corresponding architectures found by \ourmethod{} are provided in Figure~\ref{fig:explicit_encoding} and Figure~\ref{fig:NASNet_cells} for macro search space and NASNet micro search space respectively.

\subsection{Transferability}
We consider CIFAR-100 dataset \citep{cifar10} for evaluating the tranferability of the found architecture by \ourmethod{}. We use the same training setup as explained in Section~\ref{sec:extrapolation} on CIFAR-10 dataset. The training takes about 1.5 days on a single 1080Ti GPU. Results shown in Table~\ref{tab:cifar100} suggest that the learned architecture from searching on CIFAR-10 is transferable to CIFAR-100. The architecture found by \ourmethod{} achieves comparable performance to both the human-designed and RL-search generated architectures \citep{zagoruyko2016wideresnet, zhong2017blockqnn} with 10x and 2x less number of parameters respectively.

\begin{table}[!htbp]
\caption{Comparison with different classifiers on CIFAR-100.}
\label{tab:cifar100}
\centering
\scalebox{1.0}{
\begin{tabular}{c@{\hspace{2mm}}l@{\hspace{1mm}}c@{\hspace{1mm}}c@{\hspace{2mm}}c@{\hspace{1mm}}c@{\hspace{1mm}}c@{\hspace{1mm}}c c@{\hspace{2mm}} c@{\hspace{3mm}}}
\toprule
 & Architectures & {Params}  & \multicolumn{1}{c}{Test Error} & GPU & Search \\
 &  & {(M)}  & \multicolumn{1}{c}{(\%)} & Days & Method

\\ \midrule
\multirow{2}{*}{\rotatebox{90}{}}
 & Wide ResNet \citep{zagoruyko2016wideresnet} & 36.5 & 20.50  & - & manual\\
\multirow{2}{*}{\rotatebox{90}{}} 
 & Block-QNN-S \citep{zhong2017blockqnn} & 6.1 & 20.65 & 96 & RL\\
 & Block-QNN-S\footnote[1]{} \citep{zhong2017blockqnn} & 39.8 & 18.06 & 96 & RL\\
\multirow{4}{*}{\rotatebox{90}{}} 
 & \ourmethod{} & 3.3 & 20.74 & 8 & evolution\\
 & \ourmethod{}\footnote[1]{} & 11.6 & 19.83 & 8 & evolution\\
 \bottomrule
\end{tabular}}
\footnote[1]{}denotes architectures with extended number of filters
\end{table}

\subsection{Ablation Studies}
\begin{figure*}[t]
	\centering
	\begin{subfigure}[t]{.32\textwidth}
		\centering
		\includegraphics[width=0.92\textwidth]{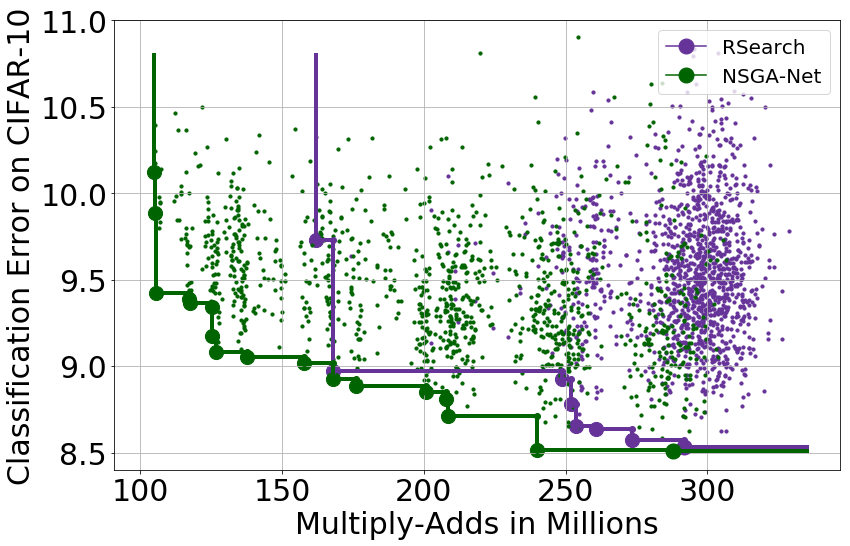}
		\caption{\label{fig:rs-frontier}}
	\end{subfigure}
	\begin{subfigure}[t]{.32\textwidth}
		\centering
		\includegraphics[width=\textwidth]{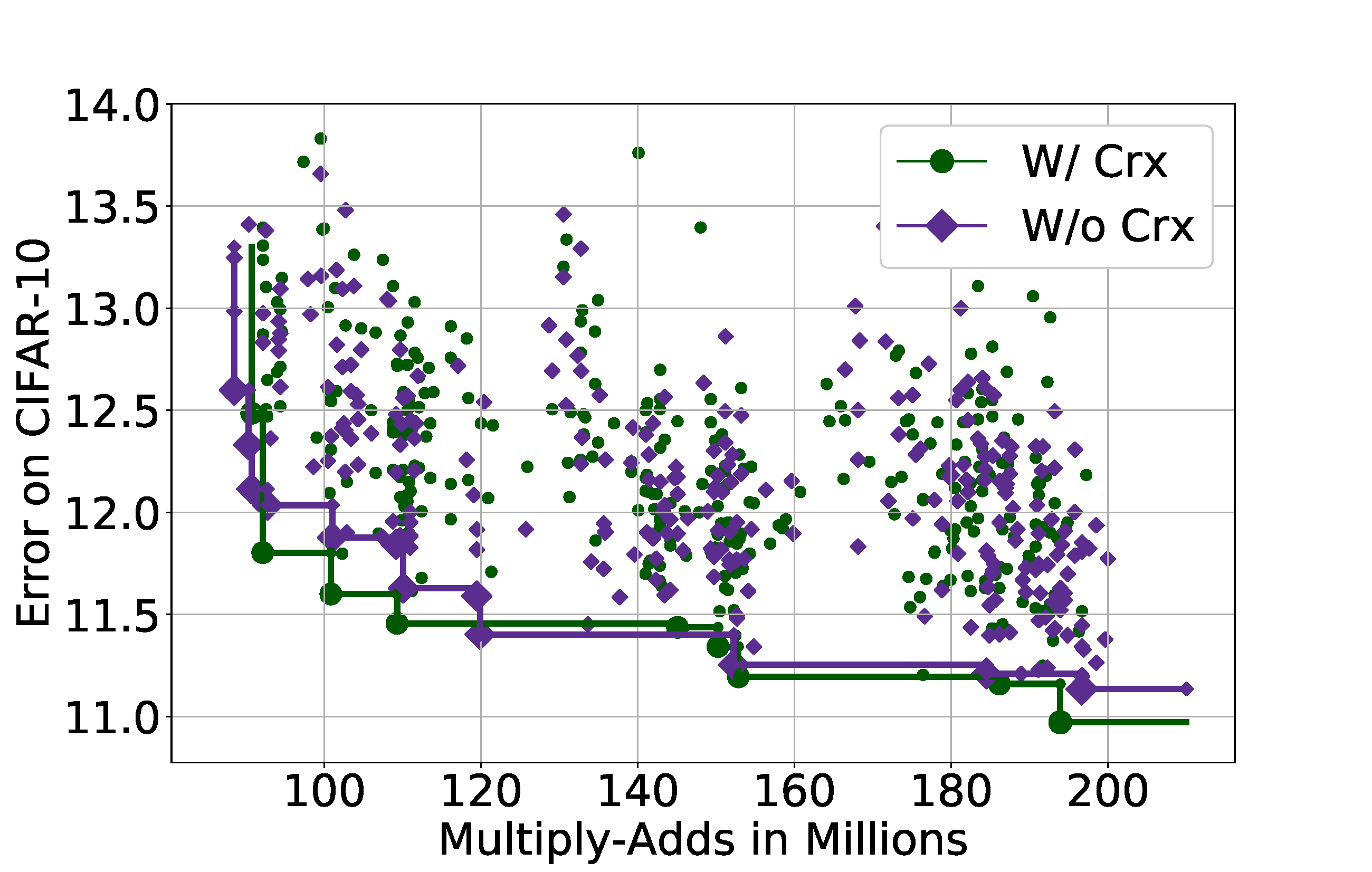}
		\caption{\label{fig:abl_crx1}}
	\end{subfigure}
	\begin{subfigure}[t]{.32\textwidth}
		\centering
		\includegraphics[width=\textwidth]{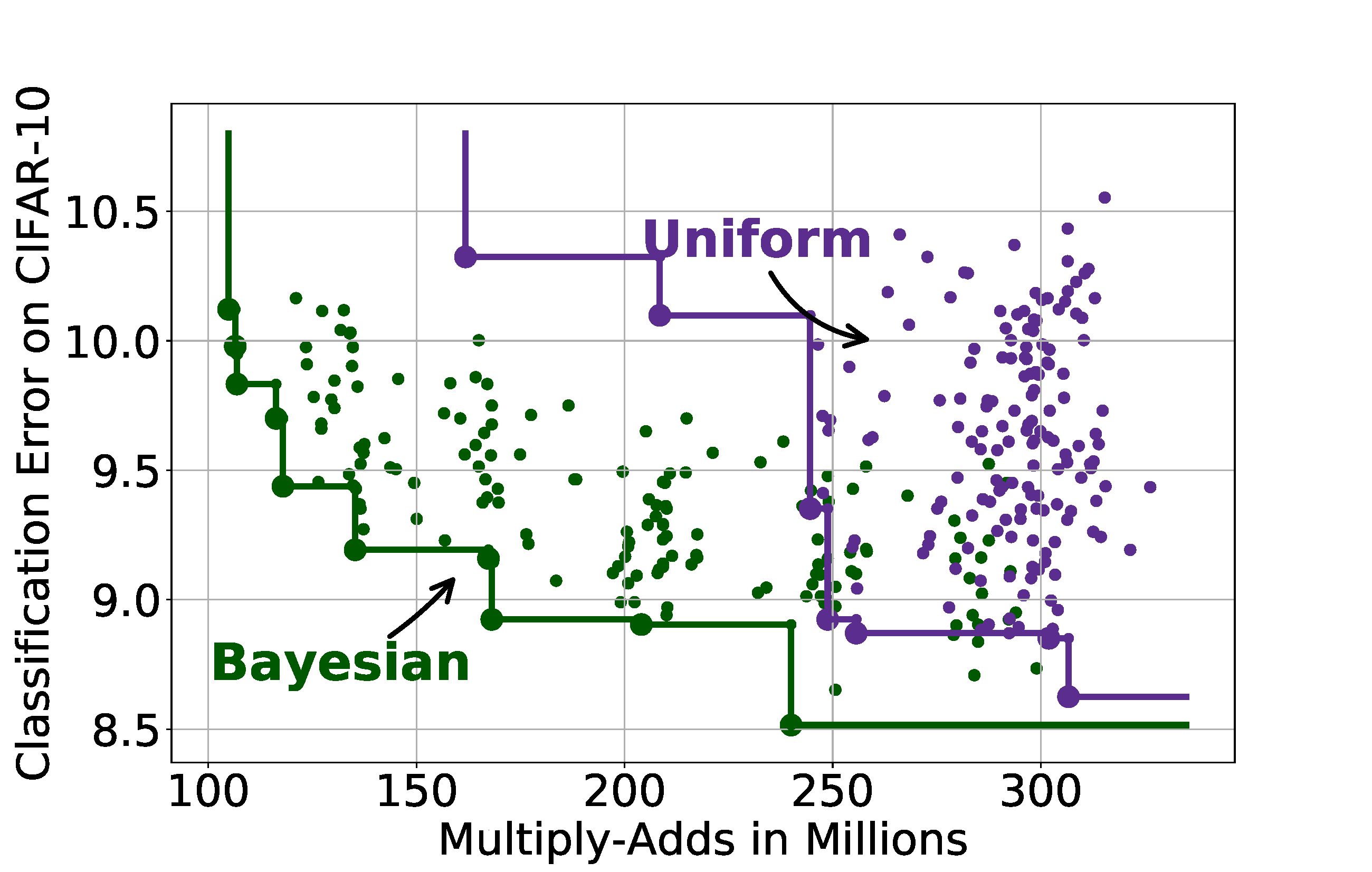}
		\caption{\label{fig:abl_bay}}
	\end{subfigure}
	\caption{(a) Trade-off frontier comparison between random search and \ourmethod{}. (b) Trade-off frontier comparison with and without crossover. (c) Comparison between sampling from uniformly from the encoding space and the Bayesian Network constructed from \ourmethod{} exploration population archive.
	\label{fig:abl}\vspace{-0.3cm}}
\end{figure*}

Here, we first present results comparing \ourmethod{} with uniform random sampling (RSearch) from our encoding as a sanity check. It's clear from Figure~\ref{fig:rs-frontier} that much better set of network architectures are obtained using \ourmethod{}. Then we present additional results to showcase the benefits of the two main components of our approach: crossover and Bayesian network based offspring creation.

\vspace{3pt}
\noindent\textbf{Crossover Operator:} Current state-of-the-art NAS search results \citep{liu2018hierarchical,real2018amoebanet} using evolutionary algorithms use mutation alone with enormous computation resources. We quantify the importance of crossover operation in an EA by conducting the following small-scale experiments on CIFAR-10. From Figure~ \ref{fig:abl_crx1}, we observe that crossover helps achieve a better trade-off frontier. 

\vspace{3pt}
\noindent\textbf{{Bayesian Network (BN) based Offspring Creation:}} Here we quantify the benefits of the exploitation stage i.e., off-spring creation by sampling from BN. We uniformly sampled 120 network architectures each from our encoding and from the BN constructed on the population archive generated by \ourmethod{} at the end of exploration. The architectures sampled from the BN dominate (see Fig.\ref{fig:abl_bay}) all network architectures created through uniform sampling.

\subsection{Discussion}
We analyze the intermediate solutions of our search and the trade-off frontiers and make some observations. Upon visualizing networks, like the one in Figure \ref{fig:explicit_encoding}, we observe that as network complexity decreases along the front, the search process gravitates towards reducing the complexity by minimizing the amount of processing at higher image resolutions i.e., remove nodes from the phases that are closest to the input to the network. As such, \ourmethod{} outputs a set of network architectures that are optimized for wide range of complexity constraints. On the other hand, approaches that search over a single repeated computational block can only control the complexity of the network by manually tuning the number of repeated blocks used. Therefore, \ourmethod{} provides a more fine-grained control over the two objectives as opposed to the control afforded by arbitrary repetition of blocks. Moreover, some objectives, for instance susceptibility to adversarial attacks, may not be easily controllable by simple repetition of blocks. A subset of networks discovered on the trade-off frontier for CIFAR-10 is provided in Appendix.

%% file: conclusion.tex
\section{Conclusions}\label{sec:conclusion}

This paper presented \ourmethod{}, a multi-objective evolutionary approach for neural architecture search. \ourmethod{} affords a number of practical benefits: (1) the design of neural network architectures that can effectively optimize and trade-off multiple competing objectives, (2) advantages afforded by population-based methods being more effective than optimizing weighted linear combination of objectives, (3) more efficient exploration and exploitation of the search space through a customized crossover scheme and leveraging the entire search history through BOA, and finally (4) output a set of solutions spanning a trade-off front in a single run. Experimentally, by optimizing both prediction performance and computational complexity \ourmethod{} finds networks that are significantly better than hand-crafted networks on both objectives and is compares favorably to other \sota{} single objective NAS methods for classification on CIFAR-10.

\section*{Acknowledgement}
This material is based in part upon work supported by the National Science Foundation under Cooperative Agreement No. DBI-0939454. Any opinions, findings, and conclusions or recommendations expressed in this material are those of the author(s) and do not necessarily reflect the views of the National Science Foundation.

%% file: appendix.tex
\section{Appendix}

\subsection{Duplicate Checking and Removal}\label{sec:duplicate}
\begin{figure}
    \centering
    \includegraphics[width=0.45\textwidth]{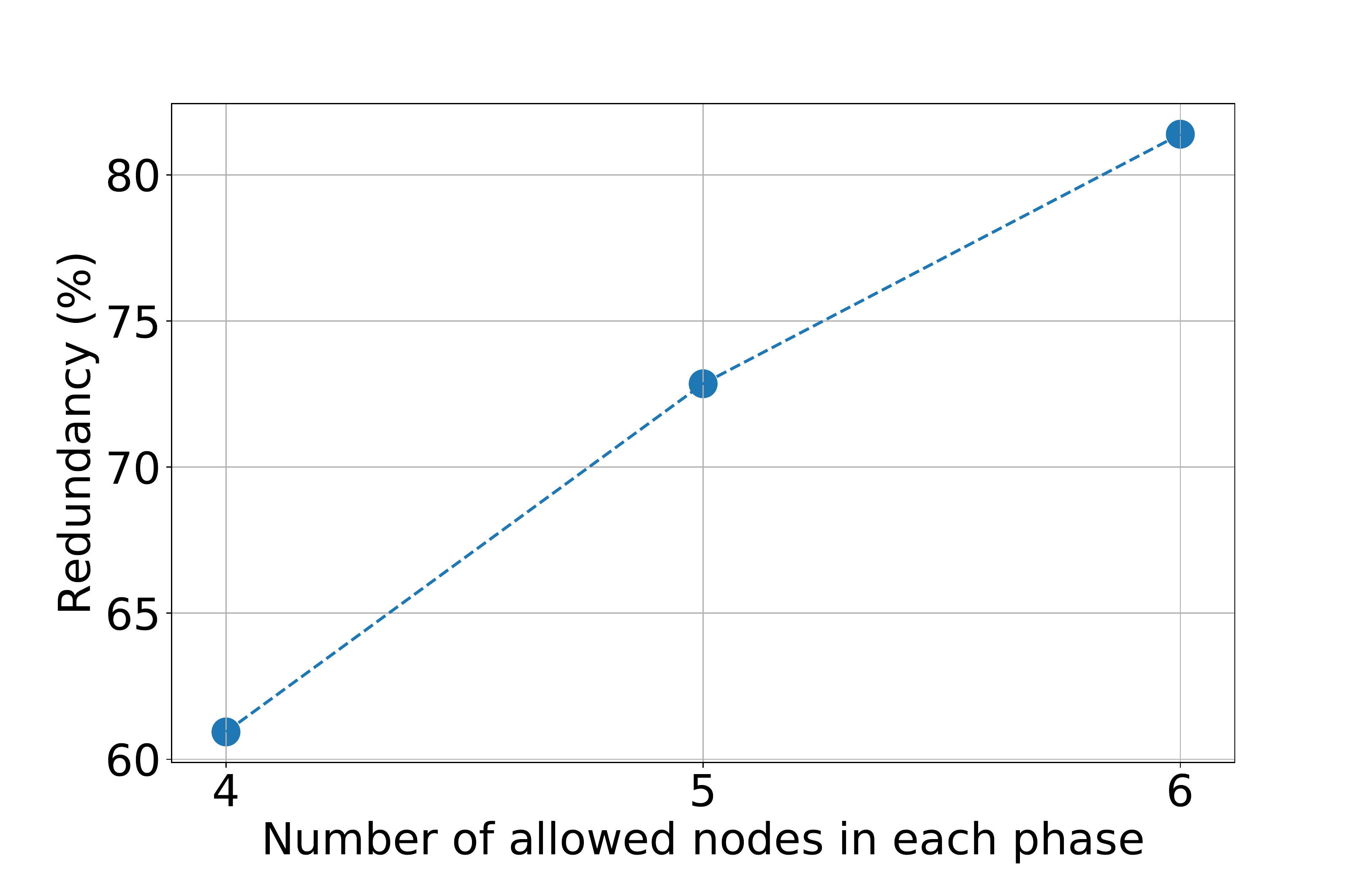}
    \caption{Increase in redundancy as node count increases.}
    \label{fig:redundancy}
\end{figure}

Due to the directed acyclic nature of our encoding, redundancy exists in the search space defined by our coding, meaning that there exist multiple encoding strings that decode to the same network architecture. Empirically, we have witnessed the redundancy becomes more and more severe as the allowed number of nodes in each phase's computational block increase, as shown in Figure~\ref{fig:redundancy}.

Since the training of a deep network is a computationally taxing task, it is essential to avoid the re-computation of the same architecture. In this section, we will provide with an overview of an algorithm we developed to quickly and approximately do a \emph{duplicate-check} on genomes. The algorithm takes two genomes to be compared as an input, and outputs a \emph{flag} to indicate if the supplied genomes decode to same architecture.

\begin{figure}
    \centering
    \includegraphics[width=0.45\textwidth]{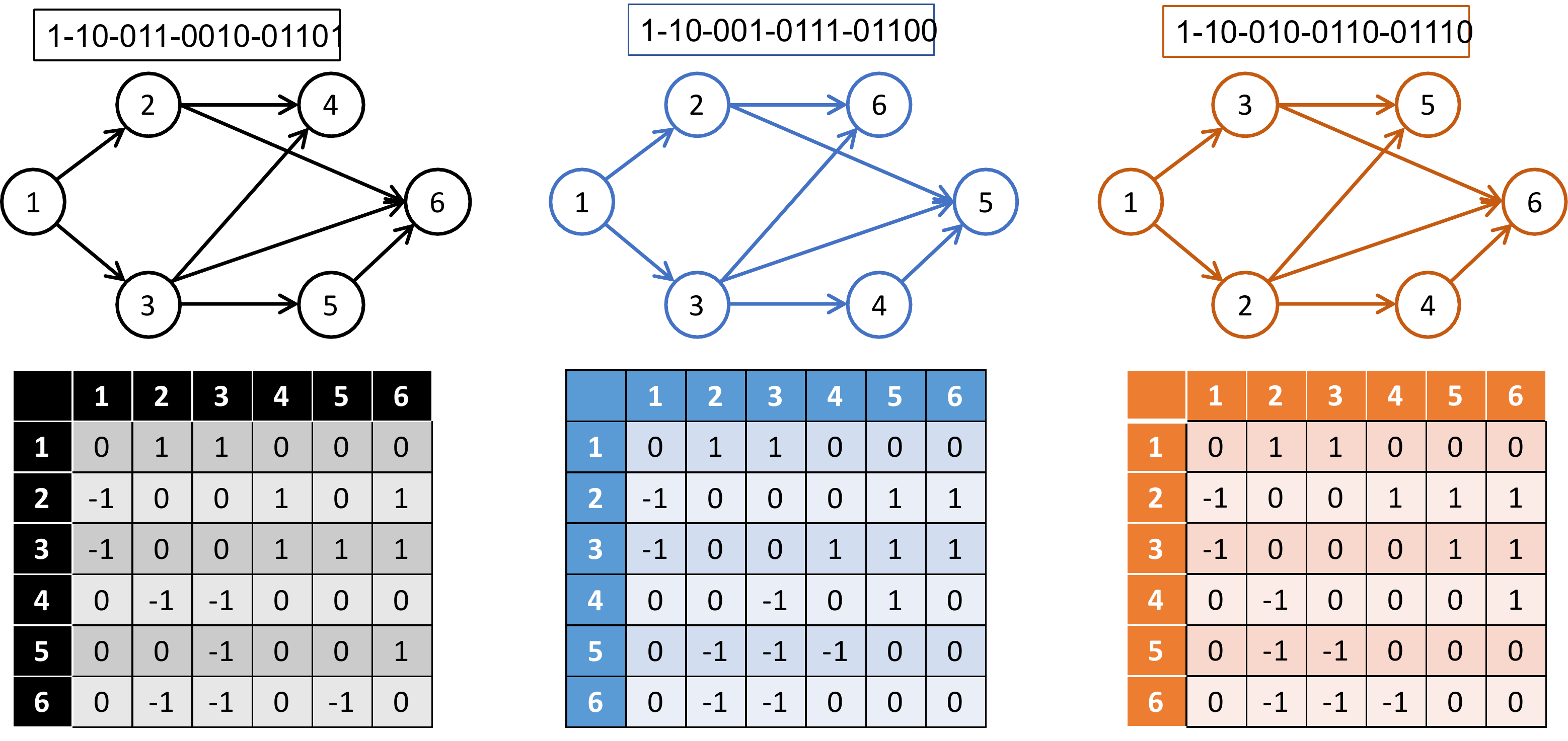}
    \caption{Examples of different encoding bit strings that decode to the same network computation block.}
    \label{fig:duplicates}
\end{figure}

In general, comparing two graphs is NP-hard, however, given that we are working with Directed Acyclic Graphs with every node being the same in terms of operations, we were able to design an efficient network architecture duplicate checking method to identify most of the duplicates if not all. The method is built on top of simply intuition that under such circumstances, the duplicate network architectures should be identified by swapping the node numbers. Examples are provided in Figure~\ref{fig:duplicates}. Our duplicates checking method first derive the connectivity matrix from the bit-string, which will have positive 1 indicating there is an input to that particular node and negative 1 indicating an output from that particular node. Then a series row-and-column swapping operation takes place, which essentially try to shuffle the node number to check if two connectivity matrix can be exactly matched. Empirically, we have found this method performs very efficiently in identifying duplicates. An example of different operation encoding bit-strings decode to the same network phase is provided in Figure~\ref{fig:duplicates}.

\subsection{Architecture Complexity Estimation}\label{sec:complexity_appendix}
We argue that the choice of inference time or number of parameters as proxies for computational complexity are sub-optimal and ineffective in practice. In fact, we initially considered both of these objectives. We concluded from extensive experimentation that inference time cannot be estimated reliably due differences and inconsistencies in computing environment, GPU manufacturer, and GPU temperature etc. Similarly, the number of parameters only relates one aspect of computational complexity. Instead, we chose to use the number of floating-point operations (FLOPs) for our second objective. The following table compares the number of active nodes, the number of connections, the total number of parameters and the FLOPs over a few sampled architecture building blocks. See Table \ref{tab:complexity} for examples of these calculations.

\begin{table*}
 \caption{Network examples comparing the number of active nodes, number of connections, number of parameters and number of multiply-adds.}\label{tab:complexity}
  \centering
  \scalebox{1.2}{
  \begin{tabular}{  m{3cm}  m{10mm}  m{10mm}  m{12mm}  m{12mm} }
    \toprule
    \multicolumn{1}{l}{Phase} & \multicolumn{1}{c}{\# of} & \multicolumn{1}{c}{\# of} & \multicolumn{1}{c}{Params.} & \multicolumn{1}{c}{FLOPs} \\ 
    \multicolumn{1}{l}{Architectures} & \multicolumn{1}{c}{Nodes} & \multicolumn{1}{c}{Conns} & \multicolumn{1}{c}{(K)} & \multicolumn{1}{c}{(M)} \\
    \midrule
    \begin{minipage}{.2\textwidth}
      \includegraphics[width=20mm, height=5mm]{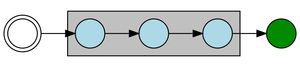}
    \end{minipage}
    & 3 & 4 & 113 & 101 \\ 
    \begin{minipage}{.2\textwidth}
      \includegraphics[width=20mm, height=7mm]{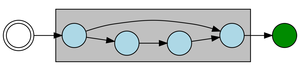}
    \end{minipage}
    & 4 & 6 & 159 & 141 \\ 
    \begin{minipage}{.2\textwidth}
      \includegraphics[width=20mm, height=12mm]{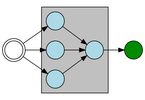}
    \end{minipage}
    & 4 & 7 & 163 & 145 \\ 
    \begin{minipage}{.2\textwidth}
      \includegraphics[width=20mm, height=10mm]{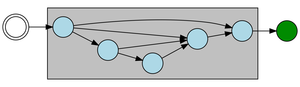}
    \end{minipage}
    & 5 & 9 & 208 & 186 \\ 
    \begin{minipage}{.2\textwidth}
      \includegraphics[width=20mm, height=10mm]{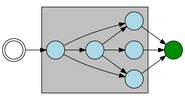}
    \end{minipage}
    & 5 & 10 & 216 & 193 \\ 
    \begin{minipage}{.2\textwidth}
      \includegraphics[width=20mm, height=10mm]{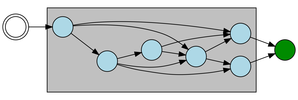}
    \end{minipage}
    & 6 & 13 & 265 & 237 \\ 
    \bottomrule
  \end{tabular}}
\end{table*}

\begin{figure*}
    \centering
    \includegraphics[width=130mm, height=220mm]{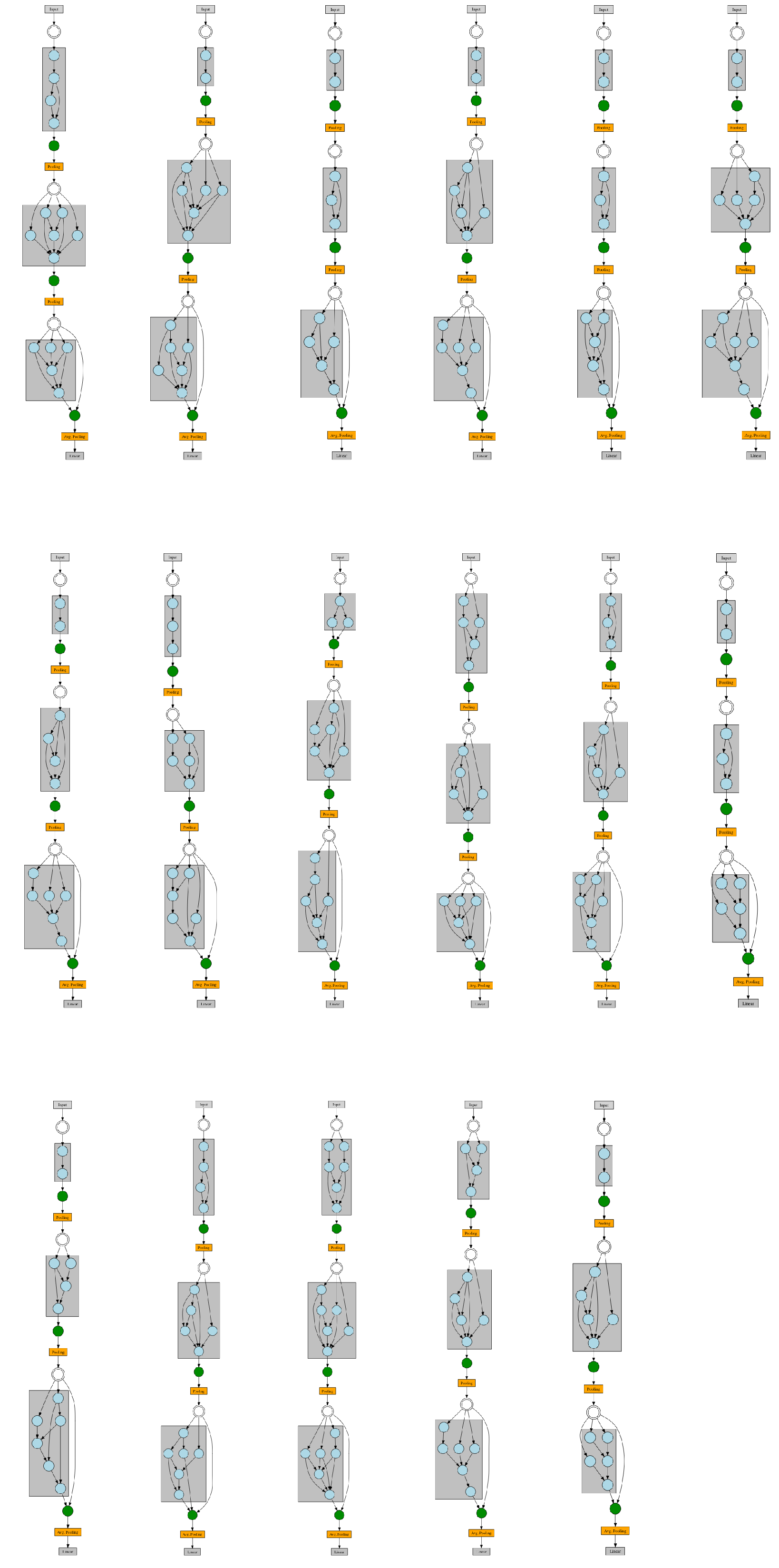}
    \caption{Set of networks architectures on the trade-off frontier discovered by \ourmethod{}.}
    \label{fig:indvs}
\end{figure*}

\begin{table*}
\caption{Summary of relevant related work along with datasets each method has been applied to, objectives optimized, and the computational power used (if reported). Methods not explicitly named are presented as the author names. PTB refers to the Penn Treebank \citep{marcus1994ptb} dataset. The Dataset(s) column describes what datasets the method performed a search with, meaning other datasets may have been presented in a study, but not used to perform architecture search. A dash represents some information not being provided. We attempt to limit the focus here to published methods, though some unpublished methods may be listed for historical contingency.}
\label{tab:related_work_summary}
\large
\centering
\scalebox{0.85}{
\begin{tabular}{l c c c c}
\toprule
 & Method Name & Dataset(s) & Objective(s) & Compute Used
\\ \midrule
\multirow{16}{*}{\rotatebox{90}{RL}} & Zoph and Lee \citep{zoph2016} & CIFAR-10, PTB & Accuracy & \tworows{800 Nvidia K80 GPUs}{22,400 GPU Hours} \\
& & & & \\
& NASNet \citep{nasnet2018} & CIFAR-10 & Accuracy & \tworows{500 Nvidia P100 GPUs}{2,000 GPU Hours} \\
& & & & \\
& BlockQNN \citep{zhong2017blockqnn} & CIFAR-10 & Accuracy & \tworows{32 Nvidia 1080Ti GPUS}{3 Days} \\
& & & & \\
& MetaQNN \citep{baker2017metaqnn} & \tworows{SVHN, MNIST}{CIFAR-10} & Accuracy & \tworows{10 Nvidia GPUs}{8-10 Days} \\ 
& & & & \\
& MONAS \citep{hsu2018monas} & CIFAR-10 & Accuracy \& Power & Nvidia 1080Ti GPUs \\ 
& & & & \\
& EAS \citep{cai2018efficient} & SVHN, CIFAR-10 & Accuracy & \tworows{5 Nvidia 1080Ti GPUs}{2 Days} \\ 
& & & & \\
& ENAS \citep{pham2018enas} & CIFAR-10, PTB & Accuracy & \tworows{1 Nvidia 1080Ti GPUs}{$<$ 16 Hours} \\ 
& & & & \\
\midrule

\multirow{24}{*}{\rotatebox{90}{EA}} & CoDeepNEAT \citep{miikulainen2017codeepneat} & CIFAR-10, PTB & Accuracy & 1 Nvidia 980 GPU \\
& & & & \\
& Real \etal \citep{real2017largescale} & CIFAR-10, CIFAR-100 & Accuracy & - \\
& & & & \\
& AmoebaNet \citep{real2018amoebanet} & CIFAR-10 & Accuracy & \tworows{450 Nvidia K40 GPUs}{\textasciitilde 7 Days} \\
& & & & \\
& GeNet \citep{genetic_cnn} & CIFAR-10 & Accuracy & \tworows{10 GPUs}{17 GPU Days} \\
& & & & \\
& NEMO \citep{kim2017nemo} & \tworows{MNIST, CIFAR-10}{Drowsiness Dataset}  & Accuracy \& Latency &  60 Nvidia Tesla M40 GPUs\\ 
& & & & \\
& Liu \etal \citep{liu2018hierarchical} & CIFAR-10 & Accuracy & 200 Nvidia P100 GPUs\\
& & & & \\
& LEMONADE \citep{elsken2018lemonade} & CIFAR-10 & Accuracy & \tworows{Titan X GPUs}{56 GPU Days} \\
& & & & \\
& PNAS \citep{liu2017progressive} & CIFAR-10 & Accuracy & - \\
& & & & \\
& PPP-Net \citep{dong2018ppp-net} & CIFAR-10 & \tworows{Accuracy \&}{Params/FLOPS/Time} & Nvidia Titan X Pascal \\
& & & & \\
\midrule

\multirow{10}{*}{\rotatebox{90}{Other}} & NASBOT \citep{kandasamy2018bayesian} & \tworows{CIFAR-10}{Various} & Accuracy & 2-4 Nvidia 980 GPUs \\
& & & & \\
& DPC \citep{chen2018randomsearch} & Cityscapes \citep{chen2014cityscapes} & Accuracy & \tworows{370 GPUs}{1 Week} \\
& & & & \\
& NAO \citep{luo2018nao} & CIFAR-10 & Accuracy & \tworows{200 Nvidia V100 GPUs}{1 Day} \\
& & & & \\
& DARTS \citep{liu2018darts} & CIFAR-10 & Accuracy & \tworows{1 Nvidia 1080Ti GPUs}{1.5 - 4 Day} \\
& & & & \\
\bottomrule
\end{tabular}}
\end{table*}